\documentclass[letterpaper]{article} 
\usepackage[a4paper, total={6in, 8in}]{geometry}
\usepackage{times}  
\usepackage{helvet}  
\usepackage{courier}  
\usepackage[hyphens]{url}  
\usepackage{graphicx} 
\usepackage{caption} 
\usepackage{algorithm}
\usepackage{algorithmic}
\usepackage{newfloat}
\usepackage{listings,xcolor}

\usepackage{soul}
\usepackage{amsmath,amssymb}
\usepackage{amsthm}
\usepackage[switch]{lineno}
\DeclareMathOperator{\E}{\mathbb{E}}

\newtheorem{example}{Example}

\newtheorem{obs}{Observation}
\newtheorem{prop}{Proposition}
\newtheorem{definition}{Definition}

\title{Quasimetric Value Functions with Dense Rewards\\}

\author{
Khadichabonu Valieva \\
University of Southern Mississippi \\
\texttt{Khadichabonu.Valieva@usm.edu} \\
\and
Bikramjit Banerjee \\
University of Southern Mississippi \\
\texttt{Bikramjit.Banerjee@usm.edu}
}

\usepackage[style=apa, backend=biber]{biblatex}
\DeclareLanguageMapping{american}{american-apa}
\newcommand{\pcite}{\parencite}  
\newcommand{\ncite}{\textcite}   

\begin{document}
  \maketitle
\begin{abstract}
As a generalization of reinforcement learning (RL) to parametrizable goals, goal conditioned RL (GCRL) has a broad range of applications, particularly in challenging tasks in robotics. Recent work has established that the optimal value function of GCRL $Q^\ast(s,a,g)$ has a quasimetric structure, leading to targetted neural architectures that respect such structure. However, the relevant analyses assume a sparse reward setting---a known aggravating factor to sample complexity. We show that the key property underpinning a quasimetric, viz., the triangle inequality, is preserved under a dense reward setting as well. Contrary to earlier findings where dense rewards were shown to be detrimental to GCRL, we identify the key condition necessary for triangle inequality. Dense reward functions that satisfy this condition can only improve, never worsen, sample complexity. This opens up opportunities to train efficient neural architectures with dense rewards, compounding their benefits to sample complexity. We evaluate this proposal in 12 standard benchmark environments in GCRL featuring challenging continuous control tasks. Our empirical results confirm that training a quasimetric value function in our dense reward setting indeed outperforms training with sparse rewards.

\end{abstract}

\section{Introduction}
Reinforcement learning (RL) is a popular class of techniques for training autonomous agents to behave (near-)optimally, often without requiring a model of the task or environment. In goal-achieving tasks, traditional RL learns policies that reach a single goal at the minimum (maximum) expected cost (value) from any state. Contrastingly in multi-task settings, a goal conditioned value function models the cost-to-go to a {\em set of goal states}, not just one. This generalization from a single-goal case to goal-conditioned RL (GCRL) yields effective representations---powered by deep neural networks---for value functions capable of capturing abstract concepts underlying goal achievement in many complex tasks~\pcite{Wang23:Optimal,Liu22:Goal-Conditioned,Plappert18:Multi-Goal}.


Recent work has established that the true optimal value function in GCRL is always a {\em quasimetric}, i.e., a metric without the constraint of being symmetric, but crucially respecting the triangle inequality~\pcite{Pitis20:Inductive,Wang22:Learning,Liu23:Metric}. This allows the search for value functions to be naturally restricted to the space of quasimetrics. Additionally, such functions are designed to be {\em universal value function approximators} (UVFA), i.e., capable of approximating arbitrarily complex value functions. Accordingly,~\ncite{Liu23:Metric} propose the metric residual network (MRN) architecture for GCRL value functions that explicitly accommodate an asymmetric component while maintaining the UVFA property and the triangle inequality. This and other similar approaches search a smaller subset of the space of value functions, yet the true optimal value function is guaranteed to reside in it. This has led to significant gains in terms of sample efficiency in recent GCRL advancements~\pcite{Liu23:Metric,Wang22:Learning,Wang23:Optimal}.

In this paper, we review some of the theoretical analyses underlying much of the work cited above. In particular, the proof of the key property of triangle inequality in~\ncite{Liu23:Metric} is established for a {\em sparse reward} setting that is easy to design but hard to learn from. By contrast, {\em dense reward} settings using various mechanisms, e.g., reward shaping, intrinsic motivation, human feedback etc., are generally known to improve sample efficiency. If dense reward-based value functions were to satisfy the triangle inequality, then their reward bias could be combined with the representational bias of quasimetrics to deliver a double punch to sample complexity. However, existing negative results ~\pcite{Plappert18:Multi-Goal} specifically in GCRL show that dense rewards significantly deteriorate the performance of state-of-the-art RL methods, and might appear to foreclose a discussion on their efficacy in GCRL. Contradictorily, we show that dense rewards can indeed bring their benefit to bear in GCRL as long as they satisfy a condition under which the triangle inequality is preserved for the {\em optimal} value function. Furthermore, we establish a condition under which the triangle inequality is preserved for {\em on-policy} value functions that may be encountered during RL iterations. This result adds  nuance to recent contradictory finding~\pcite{Wang23:Optimal} that on-policy value functions do {\em not} satisfy the triangle inequality. We show experiments in 12 benchmark GCRL tasks to establish that dense rewards indeed improve sample complexity in some tasks, but never deteriorate sample efficiency in any task. 

Our main contributions can be summarized as:
\begin{itemize}
    \item We show that using rewards shaped with potential functions that serve as admissible heuristics, the optimal value function does satisfy the triangle inequality;
    \item We define and delineate a progressive criterion for GCRL policies and show that under such policies the on-policy value function satisfies the triangle inequality;
    \item Via experiments in 12 standard benchmark GCRL tasks, we show that dense rewards improve sample complexity as well as the learned policy in 4 of the 12 tasks, while not deteriorating performance in any task.
\end{itemize}

\section{Background}
This section covers the preliminaries on goal conditioned RL, the prevalent solution approaches for GCRL, and the recent architecture of metric residual networks that we use in this paper.

\subsection{Goal-conditioned RL}
Goal conditioned RL is modeled by goal-conditioned Markov decision process, $M=({\cal S},{\cal A},{\cal G},T,R,\gamma, \rho_0,\rho_{\cal G})$. While ${\cal S},{\cal A},T,\rho_0$ define the state action spaces, the transition function and the initial state distribution just like a standard MDP, ${\cal G}$ gives the space of goal states, and $\rho_{\cal G}$ is the distribution from which a goal is sampled at the beginning of an episode. Further, the reward function $R$ is additionally parametrized by the goal, $R:{\cal S}\times {\cal A}\times {\cal G}\mapsto\Re$. In the {\em sparse reward} setting, $R$ is often defined as 
\begin{equation}
    R(s,a,g)=\begin{cases}
   0  & \text{if } M(s,a)=g\\
    -1  & \text{otherwise}
\end{cases}
\label{eqn:sparse_reward}
\end{equation}
where $M:{\cal S}\times {\cal A}\mapsto {\cal G}$ maps the product space of ${\cal S}$ and ${\cal A}$ to ${\cal G}$. As opposed to the common assumption ${\cal G}\subset {\cal S}$, $M$ allows action $a$ to decide whether the goal is reached~\pcite{Liu23:Metric}. 


\subsection{Solution Approach: DDPG+HER}
A popular approach to solving GCRL is a combination of off-policy actor-critic, e.g., DDPG~\pcite{Lillicrap16:Continuous} with hindsight experience replay (HER)~\pcite{Andrychowicz17:Hindsight}. DDPG in GCRL estimates a goal conditioned critic 
\[Q^\pi(s,a,g)=\E\left[\sum_{t=0}^{\infty}\gamma^tr_{t,g}|s_0=s,a_0=a,g\right]\]
where the expectation is taken over future steps of rewards generated by the policy ($\pi$), and the $T,R$ functions. The critic is updated by minimizing the mean squared TD error over samples $(s_t,a_t,s_{t+1},g)$ drawn from a replay buffer $D$,
\begin{equation}
    L(Q)=\E\left[(r_{t,g} +\gamma Q(s_{t+1},\pi(s_{t+1}),g)-Q(s_t,a_t,g))^2\right].
    \label{eqn:critic-loss}
\end{equation}

By ensuring that $Q$ is differentiable w.r.t. actions $a$, the actor policy $\pi$ is updated in the direction of the gradient $\E[\nabla_{a_t}Q(s_t,a_t,g)]$, where the expectation is again evaluated using samples drawn from $D$. As these samples are drawn from state distributions generated by policies different from $\pi$, DDPG is an off-policy method, although it estimates Q-values in an on-policy way (Eq.~\ref{eqn:critic-loss}). This last aspect will be scrutinized further in Sec.~\ref{sec:on-policy}.

Hindsight experience replay (HER)~\pcite{Andrychowicz17:Hindsight} mitigates the sparse reward problem by relabeling failed trajectories. 
Instead of treating all experience traces where the agent failed to achieve a goal as is, HER changes the goal in some of them to match a step of the trace in hindsight---essentially pretending as if the agent’s goal all along was to reach the state that it actually did. This transforms some of the failed episodes into successful experiences that are informative about goal achievement, and allows the agent to generalize, eventually, to the true goal distribution $\rho_{\cal G}$.

\subsection{Metric Residual Network}
~\ncite{Liu23:Metric} propose a novel neural architecture for GCRL critic based on the insight that the optimal negated action-value function, $-Q^\ast(s,a,g)$, satisfies the triangle inequality {\em in the sparse reward setting} of Eq.~\ref{eqn:sparse_reward}. Consequently, they introduce the metric residual network (MRN) that decomposes $-Q$ into the sum of a metric and an asymmetric residual component that provably approximates any quasipseudometric. Specifically, 
\begin{equation}
    Q(s,a,g) = -\left( d_{sym}(h_{sa}, h_{sg}) + d_{asym}(h_{sa}, h_{sg})\right)
    \label{eqn:Q_decomp}
\end{equation}
where $h_{sa}$ and $h_{sg}$ are latent encodings of concatenated $(s,a)$ and $(s,g)$, $d_{sym}$ and $d_{asym}$ are symmetric and asymmetric distance components given by
\begin{align}
d_{sym}(x,y) &=\|\mu_1(x)-\mu_1(y)\|, \\
d_{asym}(x,y) &=\max_i(\mu_{2i}(x)-\mu_{2i}(y))_+,
\end{align}
$\mu_1$ and $\mu_2$ are neural networks. The provable UVFA property of MRNs is due to $d_{asym}$, while $d_{sym}$ improves sample efficiency due to its symmetry. We use DDPG+HER with MRN critic architecture as the base GCRL method for this paper.

\section{Triangle Inequality}
In this section, we establish that both the optimal value function as well as on-policy value functions satisfy the triangle inequality under novel conditions.
\subsection{Optimal Value Function}
\label{sec:rs-mrn}
Our primary claim is that $-Q^\ast$ satisfies the triangle inequality not only in the sparse reward setting, but also in the presence of dense rewards, particularly potential shaped rewards. This observation lends GCRL to improved sample efficiency when approximating $-Q^\ast$ using a combination of MRN and potential shaped rewards.

We use the standard potential based shaping rewards
\begin{equation}
F(s,a,s',a',g)=\gamma\phi(s',a',g)-\phi(s,a,g)
\label{eqn:shaping_reward}
\end{equation}
and a simple potential function
\[\phi(s,a,g)=-\left(\frac{1-\gamma^{d(s,a,g)/\eta}}{1-\gamma}\right)\]
where $d$ is a distance measure between the state and the goal, and $\eta$ is a measure of the atomicity of actions---distance covered per time step. Note that in the reward regime of Eq.~\ref{eqn:sparse_reward},
\[Q^\ast(s,a,g)=-\left(\frac{1-\gamma^{L^*}}{1-\gamma}\right)\]
where $L^*$ is the {\em optimal} expected number of steps required to reach the goal $g$ from state $s$. 

\begin{obs}
If $d(s,a,g)\le \eta L^\ast$, then 
$\phi(s,a,g)\ge Q^\ast(s,a,g), \forall s,a,g$
\end{obs}
In other words, if $d(s,a,g)/\eta$ is an underestimate of $L^\ast$ then the above condition will be satisfied. Thus, $d$ acts as an admissible heuristic. For this paper, we choose a simple arc-cosine distance  $d(s,a,g)=\cos^{-1}\left( \frac{M(s,a)\cdot g}{\|M(s,a)\|\|g\|} \right)/\pi$, which is known to be a metric. Here $M$ is defined in the context of Eq.~\ref{eqn:sparse_reward}. However, this choice is not necessary for our theoretical results to hold. Rather, it is prompted by its boundedness and our desire to avoid intricate, environment-specific reward engineering.

We distinguish $Q^\ast(s,a,g)$---the optimal action values with unshaped sparse rewards---from $Q^\ast_F(s,a,g)$ which corresponds to action values with rewards shaped by $F$ in Eq.~\ref{eqn:shaping_reward}. Next we establish the validity of triangle inequality with $Q^\ast_F$ in two cases: (i) ${\cal G}\equiv {\cal S}\times{\cal A}$ and (ii) ${\cal G}\not\equiv {\cal S}\times{\cal A}$.

\subsubsection{Case I: ${\cal G}\equiv {\cal S}\times{\cal A}$}
In this setting, $M$ is the identity mapping. We use the notation $x_t=(s_t,a_t)$. The main result is:
\begin{prop}
    Consider the shaped, goal-conditioned MDP $M_{GCF}=({\cal S}, {\cal A}, {\cal G}, T, R+F, \gamma, \rho_0, \rho_g)$, with ${\cal G}\equiv {\cal S}\times{\cal A}$. The optimal universal value function $Q^\ast_F$ satisfies the triangle inequality: $\forall x^1, x^2, x^3\in {\cal X}$,
    \[Q^\ast_F(x^1,x^2)+Q^\ast_F(x^2,x^3)\le Q^\ast_F(x^1,x^3),\]
    The only condition $\phi$ must satisfy is
    \begin{equation}
        \phi(s,a,g)\ge Q^\ast(s,a,g),\forall s,a,g
        \label{eqn:assumption}
    \end{equation}
    w.r.t. the unshaped value function, for which a sufficient condition is established in Obs. 1.
\end{prop}

\noindent{\bf Proof: }As in ~\pcite{Liu23:Metric}, consider the Markov policies $\pi_1, \pi_2,\pi_3$ that are optimal w.r.t. $Q^\ast_F(x^1,x^2)$, $Q^\ast_F(x^2,x^3)$, $Q^\ast_F(x^1,x^3)$ and the (non-Markov) policy $\pi_{1\rightarrow 2}$ defined for $t>0$ as:

$\pi_{1\rightarrow 2}(a|s_t)=\begin{cases}
    \pi_1(a|s_t),  & x^2\not\in x_{<t}\\
    \pi_2(a|s_t),  & \text{otherwise.}
\end{cases}$

Let $\tau$ be the random variable that indicates the first time $\pi_{1\rightarrow 2}$ reaches $x^2$. In the steps below, we notate $F(s_t,a_t,s_{t+1},a_{t+1}, g)$ as $F_{t,g}$, and $\E_{(x_t,r_t)\sim \pi,T,R,\tau}$ as $\E_{\pi,.}$ for brevity.  Then define
\[q^1_{1\rightarrow 2}=\E_{\pi_{1\rightarrow 2},.}\bigg[ \sum\limits_{t=0}^\tau \gamma^t(r_{t,g} + F_{t,g}) | x_0=x^1,g=x^2\bigg],\]

\begin{align*}
q^2_{2\rightarrow 3}&=\E_{\pi_{1\rightarrow 2},.}\bigg[ \sum\limits_{t=\tau}^\infty \gamma^t(r_{t,g} + F_{t,g}) | x_\tau=x^2,g=x^3\bigg] \\
&=\E_{\pi_{1\rightarrow 2},.}\bigg[ \sum\limits_{t=\tau}^\infty (\gamma^tr_{t,g}) + 0 - \gamma^\tau\phi_\tau\bigg].
\end{align*}

Now,
\begin{align}
Q^\ast_F(x^1,x^2)&= \E_{\pi_1,.}\bigg[ \sum\limits_{t=0}^{\tau}\gamma^t (r_{t,g} + F_{t,g}) | x_{0}=x^1, g=x^2 \bigg]+\nonumber\\
&\E_{\pi_1,.}\bigg[ \sum\limits_{t=\tau+1}^{\infty}\gamma^t (r_{t,g} + F_{t,g}) | x_{\tau+1}=x^2, g=x^2\bigg],
\end{align}
and $\pi_1\equiv\pi_{1\rightarrow 2}$ for the first $\tau$ steps. Therefore, $Q^\ast_F(x^1,x^2)-q^1_{1\rightarrow 2}$
\begin{align*}
&= \E_{\pi_1,.}\bigg[ \sum\limits_{t=\tau+1}^{\infty}\gamma^t (r_{t,g}+ F_{t,g}) | x_{\tau}=x^2, g=x^2 \bigg],\\
&= \E_{\pi_1,.}\bigg[ \sum\limits_{t=\tau+1}^{\infty}(\gamma^t r_{t,g})+\gamma^\infty\phi(.)-\gamma^{\tau+1}\phi_{\tau+1}  | x^2,x^2\bigg]\\
&= \E_{\pi_1,.}\bigg[ \sum\limits_{t=\tau+1}^{\infty}(\gamma^t r_{t,g})-\gamma^{\tau+1}\phi_{\tau+1} | 
x^2,x^2
\bigg] \\
&= \E_{\tau}[\gamma^{\tau+1}][Q^\ast(x^2, x^2)-\phi(x^2, x^2)]\\
&\le 0 \text{ by assumption (Eq.~\ref{eqn:assumption}).}
\end{align*}
Similarly, 
\begin{align}
    Q^\ast_F(x^2,x^3) &= \E_{\pi_2,.}\bigg[ \sum\limits_{t=0}^{\infty}\gamma^t (r_{t,g}+F_{t,g}) | x_{0}=x^2, g=x^3 \bigg]\nonumber\\
    &\le \gamma^\tau \E_{\pi_2,.}\bigg[ \sum\limits_{t=0}^{\infty}\gamma^t (r_{t,g}+F_{t,g}) | x^2,x^3 
    \bigg]\nonumber\\
    &=\E_{\pi_2,.}\bigg[ \sum\limits_{t=0}^{\infty}\gamma^{t+\tau} (r_{t,g}+F_{t,g}) | x^2,x^3 
    \bigg]\nonumber\\
    &=\E_{\pi_2,.}\bigg[ \sum\limits_{k=\tau}^{\infty}\gamma^{k} r_{k,g} +0-\gamma^\tau\phi_\tau | x^2,x^3 
    \bigg]\nonumber\\
    &=\E_{\pi_{1\rightarrow 2},.}\bigg[ \sum\limits_{k=\tau}^{\infty}\gamma^{k} r_{k,g} -\gamma^{\tau}\phi_\tau| x^2,x^3 
    \bigg]\nonumber\\
    &=q^2_{2\rightarrow 3},
\end{align}
since $\pi_2\equiv\pi_{1\rightarrow 2}$ after $\tau$. Therefore, $Q^\ast_F(x^2,x^3)-q^2_{2\rightarrow 3}\le 0$. Consequently, 
\[Q^{\pi_{1\rightarrow 2}}_F(x^1,x^3)=q^1_{1\rightarrow 2}+q^2_{2\rightarrow 3}\ge Q^\ast_F(x^1,x^2)+Q^\ast_F(x^2,x^3).\]
But since the optimal $Q^\ast_F(x^1,x^3)\ge Q^{\pi_{1\rightarrow 2}}_F(x^1,x^3)$, we arrive at the triangle inequality. \qed


\subsubsection{Case II: ${\cal G}\not\equiv {\cal S}\times{\cal A}$}
The proof of this case closely resembles~\pcite{Liu23:Metric}; we highlight the main difference in \textcolor{blue}{blue} color but also provide the rest of the proof for completeness. In this case, $M$ is an onto mapping. Given a goal $g$, the GCRL problem effectively reduces to a standard MDP and there exists a deterministic optimal policy $\pi^\ast$ for reaching the goal $g$ from an initial state $x=(s_0,a_0)$. Then, under deterministic dynamics,
\[Q^\ast_F(x,g)=\sup_{x':M(x')=g}Q^\ast_F(x,x').\]
Assuming the supremum is attainable, let
\begin{equation}
x_g = \arg \max_{x':M(x')=g} Q^\ast_F(x, x'),
\label{eqn:x_g}
\end{equation}
then $Q^\ast_F(x, g) = Q^\ast_F(x, x_g)$. Assume for contradiction that this is not the case, i.e., $Q^\ast_F(x, g) \neq Q^\ast_F(x, x_g)$. There are two possibilities:

\begin{itemize}
    \item {If \(Q^\ast_F(x, x_g) > Q^\ast_F(x, g)\):} This would imply that by using a policy that selects $x_g$ rather than $g$, one could achieve a higher return. This contradicts the definition of $\pi^\ast$ as the optimal policy, thus $Q^\ast_F(x, x_g) > Q^\ast_F(x, g)$ cannot be true.
    \item {If \(Q^\ast_F(x, x_g) < Q^\ast_F(x, g)\):} Let
    \[
    \tau = \min_t (M(x_t) = g), 
    \]
    such that $x_\tau$ is the first $(s, a)$ pair along the optimal trajectory that achieves the goal.
There are two further cases:
\begin{enumerate}
    \item {After reaching $x_\tau$, $\pi^\ast$ will repeatedly return to $x_\tau$.} In this case, we have $Q^\ast_F(x,x_g) \ge Q^\ast_F(x,x_\tau)$ by the definition of $x_g$ (Eq.~\ref{eqn:x_g}) and
    \begingroup
    \color{blue}{
    \begin{align}
    Q^\ast_F(x,x_\tau) &= Q^\ast_F(x,g) - \gamma^{\tau+1} Q^\ast_F(x_\tau, g) \nonumber\\
    &= Q^\ast_F(x,g) - \gamma^{\tau+1} \left[ Q^\ast(x_\tau, g) - \phi(x_\tau, g) \right] \nonumber\\
    &\ge Q^\ast_F(x,g), \text{ by Eq.~\ref{eqn:assumption}}.
    \label{eqn:case_2}
    \end{align}}\endgroup
    Combining the two, we get $Q^\ast_F(x,x_g)\ge Q^\ast_F(x,g)$
    which contradicts our assumption that $Q^\ast_F(x, g) > Q^\ast_F(x, x_g)$. 

    \item {$\pi^\ast$ never returns to $x_\tau$ after reaching it for the first time.} In this case, one can find the next $\tau' = \min_{t>\tau} (M(x_t) = g), 
    $ such that $x_{\tau'}$ is another $(s,a)$ along the optimal trajectory. Again,  there are two sub-cases:
    \begin{enumerate}
        \item If $\pi^\ast$ repeatedly visits $x_{\tau'}$, then the argument in the first case applies. 
        \item Otherwise, recursively find the next $\tau''$, and so on. Eventually, we may have a  last state $x_\zeta$ such that no $t > \zeta$ satisfies $M(x_t) = g$. Then,
        $
        Q^\ast_F(x, x_g) \ge $ 
        \begingroup\color{blue}{$Q^\ast_F(x, x_\zeta) \ge Q^\ast_F(x, g).
        $}
    The last inequality is derived in the same way as Eq.~\ref{eqn:case_2}. ~\endgroup 
 Alternatively, there may exist an infinite sequence of such $\{x_\tau\}$. Following this sequence, the claim remains true but the supremum is not attainable. However, in this case an $x_\tau$ can be found in the sequence such that $Q^\ast_F(x, x_\tau)$ is arbitrarily close to $Q^\ast_F(x, g)$.\qed
    \end{enumerate}
\end{enumerate}
\end{itemize}

\subsubsection{Projection}
\label{sec:projection}
$Q_F^\ast$ has the same upper bound as $Q^\ast$, since $Q_F^\ast(s,a,g)=Q^\ast(s,a,g)-\phi(s,a,g)\le 0$ by Eq.~\ref{eqn:assumption}. Consequently, the MRN architecture needs no modification, specifically to Eq.~\ref{eqn:Q_decomp}, as the critic output is guaranteed to be non-positive despite potentially positive shaping rewards. However, $Q_F^\ast$ has a more informed lower bound:
\begin{align}
    Q^\ast_F(s,a,g) &= Q^\ast(s,a,g)-\phi(s,a,g) \nonumber\\
    &\ge -\frac{1}{1-\gamma} -\phi(s,a,g) \nonumber\\
    &=-\frac{\gamma^{d(s,a,g)/\eta}}{1-\gamma}
    \label{eqn:projection}
\end{align}
which we impose on the critic. Recent analyses~\pcite{Gupta22:Unpacking} have shown that projection informed by shaping effectively reduces the size of the state space for exploration, leading to improved regret bounds.

\subsection{On-Policy Value Functions}
\label{sec:on-policy}
In their critique of on-policy Q-function estimation methods for GCRL such as DDPG in continuous control tasks, ~\ncite{Wang23:Optimal} show that {\em on-policy} Q-function may not be a quasimetric, even though the {\em optimal} Q-function is. However, their counterexample is an extreme policy that is unlikely to be encountered during on-policy iterations. In this section, we establish that on-policy Q-functions do indeed satisfy the triangle inequality (and hence meet the quasimetric criterion) if the policy makes a minimal progress toward the goal. We call such policies {\em progressive policies} and believe they are more relevant to on-policy Q-function estimation in GCRL. We first formalize the notion of progressive policies, specify our assumption, and finally show that the corresponding value functions satisfy the triangle inequality.

For notational convenience, we write $\E_{s'\sim T(.|s,a),a'\sim\pi(s')}$ simply as $\E_{s',a'}$. Note that the on-policy value function for a policy $\pi$ satisfies
\begin{equation}
    Q^\pi(s,a,g)=R(s,a,g)+\gamma \E_{s',a'}\bigg[Q^\pi(s',a',g)\bigg].
    \label{eqn:on-policy-value-fn}
\end{equation}
\begin{definition}
    The progress of a GCRL policy $\pi$ is given by
    \[\Delta^\pi(s,a,g)=\E_{s',a'}\bigg[ Q^\pi(s',a',g)\bigg]-Q^\pi(s,a,g)\]
    for any $(s,a,g)\in {\cal S}\times {\cal A}\times {\cal G}$.
    \label{def:progressive-policy}
\end{definition}

We refer to $\Delta^\pi$ for the optimal policy as $\Delta^\ast$. We assume that the progress of $\pi$ is not unboundedly different from that of the optimal policy, i.e., the following holds for some $0<\epsilon<\infty$
\begin{equation}
    \epsilon \le \Delta^\ast(s,a,g)-\Delta^\pi(s,a,g)\le 2\epsilon.
    \label{eqn:progress-bound}
\end{equation}
Note that (i) $\epsilon$ does not need to be small, just finite; (ii) the counterexample in~\ncite{Wang23:Optimal} does not satisfy this assumption. Our main result of this section is:

\begin{prop}
Consider the goal-conditioned MDP $M_{GC}=({\cal S}, {\cal A}, {\cal G}, T, R, \gamma, \rho_0, \rho_g)$. The on-policy value function $Q^\pi$ defined in Eq.~\ref{eqn:on-policy-value-fn} for any policy $\pi$ that satisfies Eq.~\ref{eqn:progress-bound} also satisfies the triangle inequality: $\forall x^1, x^2, x^3\in {\cal X}$,
    \[Q^\pi(x^1,x^2)+Q^\pi(x^2,x^3)\le Q^\pi(x^1,x^3).\]
\end{prop}
\noindent{\bf Proof:} From Eq.~\ref{eqn:on-policy-value-fn} we have,
\[\E_{s',a'}\bigg[Q^\pi(s',a',g)\bigg] = (Q^\pi(s,a,g)-R(s,a,g))/\gamma.\]
Then, using Eq.~\ref{eqn:progress-bound} and Def.~\ref{def:progressive-policy}, for either $z\equiv (x^1,x^2)$ or $z\equiv (x^2,x^3)$, the following holds: 
\begin{align*}
    \Delta^\ast(z)-\Delta^\pi(z) &= \frac{Q^\ast(z)-R(z)}{\gamma}-Q^\ast(z) -\\
    &\frac{Q^\pi(z)-R(z)}{\gamma}+Q^\pi(z)\\
    &=(\frac{1}{\gamma}-1)[Q^\ast(z)-Q^\pi(z)]\\
    &\ge \epsilon ~\text{ (by Eq.~\ref{eqn:progress-bound}).}
\end{align*}
Adding for $z\equiv (x^1,x^2)$ and $z\equiv (x^2,x^3)$, we get
\begin{equation}
    Q^\pi(x^1,x^2)+Q^\pi(x^2,x^3)\le Q^\ast(x^1,x^2)+Q^\ast(x^2,x^3)-\frac{2\epsilon\gamma}{1-\gamma}
    \label{eqn:triangle-LHS}
\end{equation}
But similarly for $z\equiv (x^1,x^3)$,
    \[\Delta^\ast(z)-\Delta^\pi(z) = (\frac{1}{\gamma}-1)[Q^\ast(z)-Q^\pi(z)]
    \le 2\epsilon\] 
    by Eq.~\ref{eqn:progress-bound}. This gives $Q^\ast(x^1,x^3)\le Q^\pi(x^1,x^3)+\frac{2\epsilon\gamma}{1-\gamma}$. Finally, the result is obtained by combining this with Eq.~\ref{eqn:triangle-LHS} and noting that the triangle inequality holds for the optimal Q-value function, i.e., $Q^\ast(x^1,x^2)+Q^\ast(x^2,x^3)\le Q^\ast(x^1,x^3)$.\qed
    
This result relies on the triangle inequality of the optimal value function as established before in ~\pcite{Liu23:Metric} for sparse rewards and in Sec.~\ref{sec:rs-mrn} for dense rewards. But it does not have any dependence on whether $M$ is one-to-one or onto, hence the two cases ${\cal G}\equiv {\cal S}\times {\cal A}$ and ${\cal G}\not\equiv {\cal S}\times {\cal A}$ do not need to be distinguished. The result also does not assume any specific form of, or bounds on, the reward function. Hence it extends readily to shaped rewards as well, as long as the shaped value function respects the same upper bound (Sec.~\ref{sec:projection}), $Q^\pi_F(.)\le 0$.

\section{Related Work}
Several value function representations have been proposed for GCRL over the last decade.~\ncite{Schaul15:Universal} introduced the bilinear decomposition, later generalized to bilinear value networks~\pcite{Yang22:Bilinear} with better learning efficiency.~\ncite{Pitis20:Inductive} proposed the deep norm (DN) and wide norm (WN) families of neural representations that respect the triangle inequality. However, they are restricted to norm-induced functions, and are generally unable to represent all functions that respect the triangle inequality. By contrast, Possion Quasi-metric Embedding (PQE)~\pcite{Wang22:Learning} can universally approximate {\em any} quasipseudometric, thus improving upon DN/WN. However, as~\ncite{Liu23:Metric} argue, PQE captures the restrictive form of first hitting-time when applied to GCRL, whereas MRNs capture the more general setting of repeated return to goal ($Q^\ast(g,g)\neq 0$), while preserving the UVFA property of PQEs.~\ncite{Durugkar21:Adversarial} introduced a quasimetric that estimates the Wasserstein-1 distance between state visitation distributions, minimizing which is equivalent to policy optimization in GCRL tasks with deterministic transition dynamics. While they use the Wasserstein discriminator as a potential for reward shaping (as intrinsic motivation), our goal is different. We prove that dense rewards via shaping preserves the triangle inequality for the general class of potential based shaping, not just for the Wasserstein based quasimetric. Other recent architectures for GCRL use contrastive representation~\pcite{Eysenbach22:Contrastive} but without regard to quasimetric architecture, and Quasimetric RL (QRL)~\pcite{Wang23:Optimal} where temporal distances are learned, although it is unclear if it respects the triangle inequality in stochastic settings.

While the above literature on representation learning has been centered on expressive and flexible representations for GCRL, their analyses are generally restricted to sparse reward settings. 
In fact, past experimentation with dense rewards in GCRL have yielded negative results~\pcite{Plappert18:Multi-Goal}.~\ncite{Plappert18:Multi-Goal} argue that 
dense reward signals are hard to learn from because (i) arbitrary distance measures (e.g., Euclidean distance and quaternions for rotations) are highly non-linear; (ii) dense rewards bias the policy toward specific strategies that may be sub-optimal. Similar arguments also appear in~\pcite{Liu22:Goal-Conditioned}. However, our setting overcomes these objections. First, we establish the sufficient condition (Eq.~\ref{eqn:assumption}) for the triangle inequality that may not be satisfied by arbitrary distance measures, $\phi$, providing guidance on the contrary. And second, we use potential based reward shaping~\pcite{Ng99:Policy} which is policy invariant, hence strategically unbiased. However, we acknowledge the large body of work on reward shaping~\pcite{Tang17:Exploration,Brys14:Combining,Devlin12:Dynamic,Knox09:Interactively,VanSeijen17:Hybrid} of various types (e.g., count-based, intrinsic motivation, human advice, etc.) where careful, heuristic reward design is often employed to explicitly bias the policies.

\begin{figure}
    \centering
    \includegraphics[width=\linewidth]{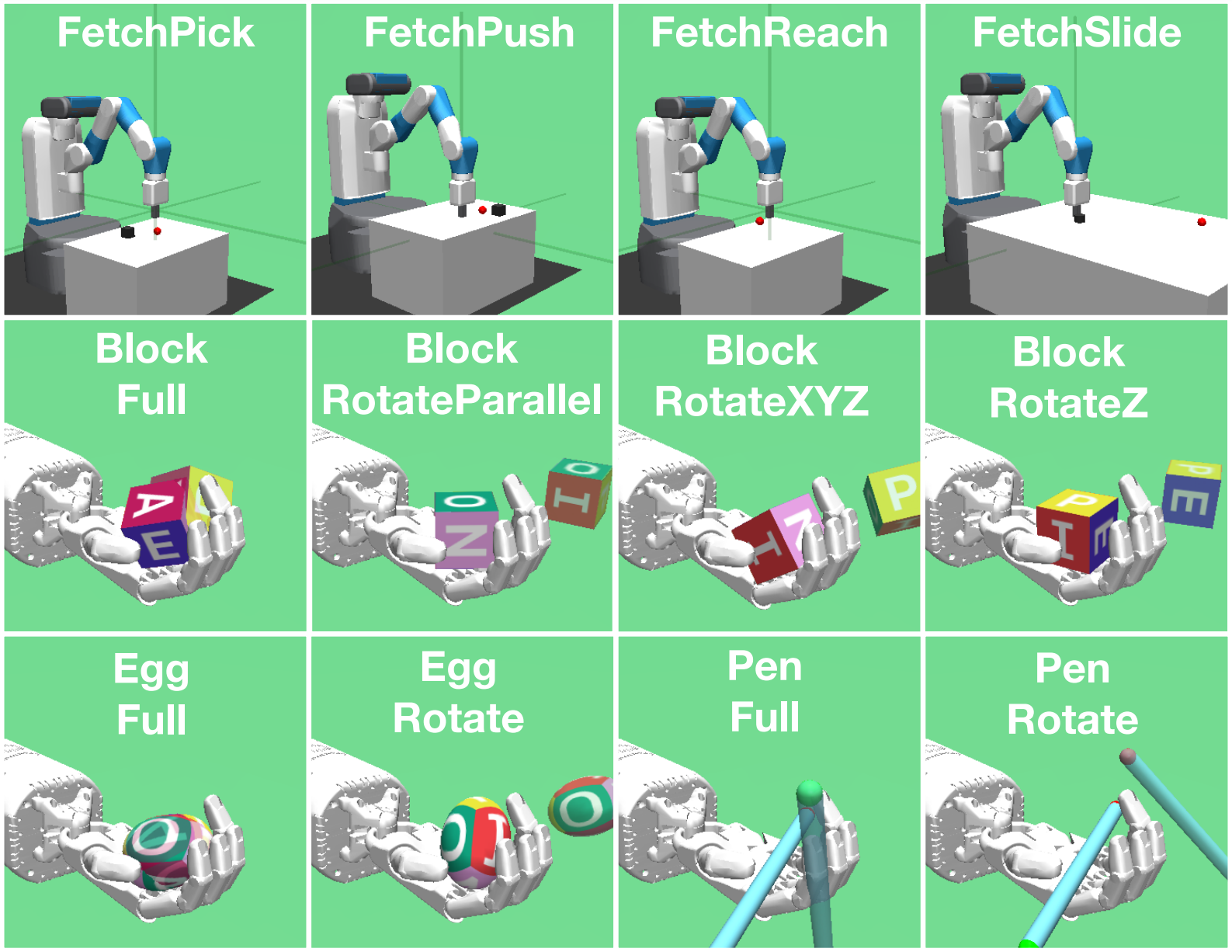}
    \caption{GCRL benchmark environments~\pcite{Plappert18:Multi-Goal}. Figure from~\pcite{Liu23:Metric}.}
    \label{fig:domain}
\end{figure}

\begin{figure*}[tbh]
    \centering
    \includegraphics[width=\linewidth]{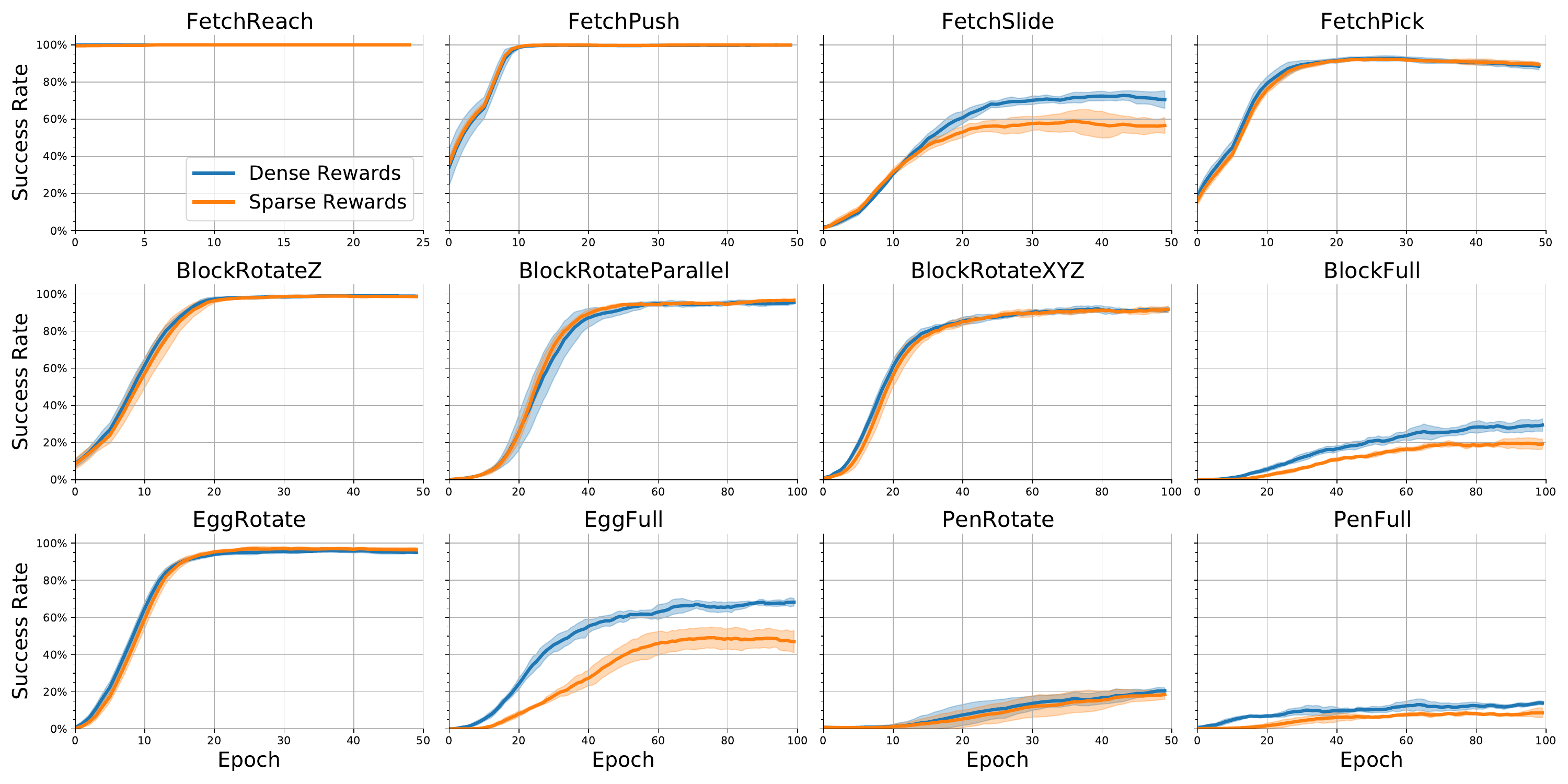}
    \caption{Comparison of MRN with sparse rewards vs. dense rewards. Learning curves are averaged over five independent trials, and one standard deviation bands are included. We see statistically significant improvement of performance due to dense rewards in 4 of the 12 environments, viz., FetchSlide, BlockFull, Eggfull and PenFull. There is no statistically significant deterioration in any environment.}
    \label{fig:plots}
\end{figure*}
\section{Experimental Results}

We use GCRL benchmark manipulation tasks with the Fetch robot and Shadow-hand domains~\pcite{Plappert18:Multi-Goal}; see Fig.~\ref{fig:domain}. MRN has been extensively compared with competitive baseline architectures and found to be superior, viz., BVN~\pcite{Yang22:Bilinear}, DN/WN~\pcite{Pitis20:Inductive}, and PQE~\pcite{Wang22:Learning}. Consequently, we focus on comparing against MRN with sparse rewards as the sole baseline. 

We experimentally evaluate the following hypotheses:

{\bf Hypothesis 1:} Dense rewards can be used in conjunction with MRN architecture for estimating value functions. Specifically, the property of $Q^\ast$ function that MRNs capture---that it satisfies the triangle inequality---is preserved in the presence of shaped rewards with the new value function $Q_F^\ast$. Dense rewards enable the less restrictive $Q_F^\ast$ to be learned more efficiently than $Q^\ast$.

{\bf Hypothesis 2:} Plappert et al.~\pcite{Plappert18:Multi-Goal} found that dense rewards hurt RL performance in GCRL robot manipulation tasks. This negative result contradicts our Hypothesis 1. We conjecture that their application of dense rewards did not satisfy the required structure---specifically Eq.~\ref{eqn:assumption}---which is why it failed. To confirm this contradiction, we verify that our dense reward setting does not deteriorate RL performance in any task. 

We use the MRN code publicly available at:~\texttt{\url{https://github.com/Cranial-XIX/metric-residual-network}} with simple modifications to add Eq.~\ref{eqn:shaping_reward} to the reward function and Eq.~\ref{eqn:projection} to clip the critic's output. No other changes were made to any algorithm or neural architecture. In particular, all parameter values (e.g. layer sizes) were unchanged, except the newly added parameter $\eta$ was set to $0.02$. This value was selected from the set $\{0.01,0.02,0.03,0.04,0.05\}$ using performance improvement as the criterion. 
For each environment, 5 seeds were used for independent trials, as in~\pcite{Liu23:Metric}. In each epoch, the agent is trained on 1000 episodes and then evaluated over 100 independent rollouts with randomly sampled goals. The average success rates in these evaluations are collected over 5 seeds. The results are plotted in Fig.~\ref{fig:plots}. All experiments were run on NVIDIA Quadro RTX 6000 GPUs with 24 GiB of memory each and running on Ubuntu 22.04. 

We see from Fig.~\ref{fig:plots} that indeed dense rewards improve the sample complexity in some environments, to an extent that is statistically significant as shown with standard deviation bands. In particular, there is statistically significant improvement in 4 of the 12 environments, viz., FetchSlide, BlockFull, Eggfull and PenFull. Not only is the sample complexity improved, but also higher quality policies are learned. This confirms Hypothesis 1. Furthermore, no statistically significant deterioration is observed in any environment, confirming Hypothesis 2.

\section{Conclusion}
We have presented generalizations of previous results on triangle inequality in the context of value functions in GCRL. Specifically, we have shown that the optimal value function satisfies the triangle inequality even when the reward function is densified with a particular class of shaping functions. Additionally, we have shown that the on-policy value functions also satisfy the triangle inequality if the underlying policy satisfies a certain progressive criterion. Both of these findings contradict previously published results in some ways, which emphasizes the importance of the nuanced conditions behind our results. Experiments in 12 benchmark GCRL tasks confirms that dense rewards only improve the sample efficiency, never deteriorates it. Future investigations could focus on more general classes of reward functions that preserve the quasimetric property of value functions and/or lend themselves to other, potentially more effective, architectures.

\printbibliography
\end{document}